\documentclass[letterpaper, 10 pt, conference]{ieeeconf}  

\IEEEoverridecommandlockouts                              

\usepackage{microtype}
\usepackage{cite}
\usepackage{amsmath,amsfonts}
\usepackage[normalem]{ulem}
\usepackage{xpatch}
\usepackage{graphicx}
\usepackage{xcolor}
\usepackage{multirow}
\usepackage{colortbl}
\usepackage{makecell}
\usepackage{caption}
\usepackage{url}
\usepackage{multirow}
\usepackage{subcaption}
\usepackage{hyperref}
\usepackage[absolute,overlay]{textpos}
\title{\LARGE \bf
Design and Validation of a Wireless Drone Docking Station
}

\author{Dario Stuhne, Goran Vasiljević, Stjepan Bogdan and Zdenko Kovačić
\thanks{*This work is supported by: I. the project AERIAL COgnitive Integrated Multi-task Robotic System with Extended Operation Range and Safety (AERIAL CORE) funded by European Union’s Horizon 2020 Research and Innovation Programme under Grant 871479, and II. the European Union's Horizon Europe research program Widening participation and spreading excellence, through project Strengthening Research and Innovation Excellence in Autonomous Aerial Systems (AeroSTREAM) - Grant agreement ID: 101071270.}
\thanks{Authors are with LARICS Laboratory for Robotics and Intelligent Control Systems, University of Zagreb, Faculty of Electrical Engineering and Computing, Unska 3, 10000 Zagreb, Croatia
        {\tt\small {dario.stuhne, goran.vasiljevic, stjepan.bogdan, zdenko.kovacic}@fer.hr}}%
}

\begin{document}

\maketitle
\thispagestyle{empty}
\pagestyle{empty}
\maxdeadcycles=20000
\begin{textblock*}{14.9cm}(3.2cm,0.75cm) %
	{\footnotesize © 2023 IEEE.  Personal use of this material is permitted.  Permission from IEEE must be obtained for all other uses, in any current or future media, including reprinting/republishing this material for advertising or promotional purposes, creating new collective works, for resale or redistribution to servers or lists, or reuse of any copyrighted component of this work in other works.}
\end{textblock*}
\begin{abstract}

Drones are increasingly operating autonomously, and the need for extending drone power autonomy is rapidly increasing. One of the most promising solutions to extend drone power autonomy is the use of docking stations to support both landing and recharging of the drone. To this end, we introduce a novel wireless drone docking station with three commercial wireless charging modules. We have developed two independent units, both in mechanical and electrical aspects: the energy transmitting unit and the energy receiving unit. We have also studied the efficiency of wireless power transfer and demonstrated the advantages of connecting three receiver modules connected in series and parallel. We have achieved maximum output power of 96.5 W with a power transfer efficiency of 56.6\% for the series connection of coils. Finally, we implemented the system in practice on a drone and tested both energy transfer and landing.

\end{abstract}

\section{Introduction}
\label{sec:intro}

The drone market is growing rapidly. The market is expected to reach nearly US\$100 billion by 2030, a fivefold increase from 2020 \cite{statistaGlobalDrone}. Huge growth in drone adoption is forecast in the logistics, industrial, and agricultural sectors \cite{mourgelas2020} for a variety of tasks such as monitoring and inspection of goods and assets, video surveillance, delivery, environmental mapping, etc. Regardless of the type of sector in which the drone is used, it is expected to be operational 24 hours a day in the future. To ensure this, the drone's energy demand and consumption should be taken into account and handled appropriately.

\begin{figure}[h]
\centering
\includegraphics[width=\linewidth]{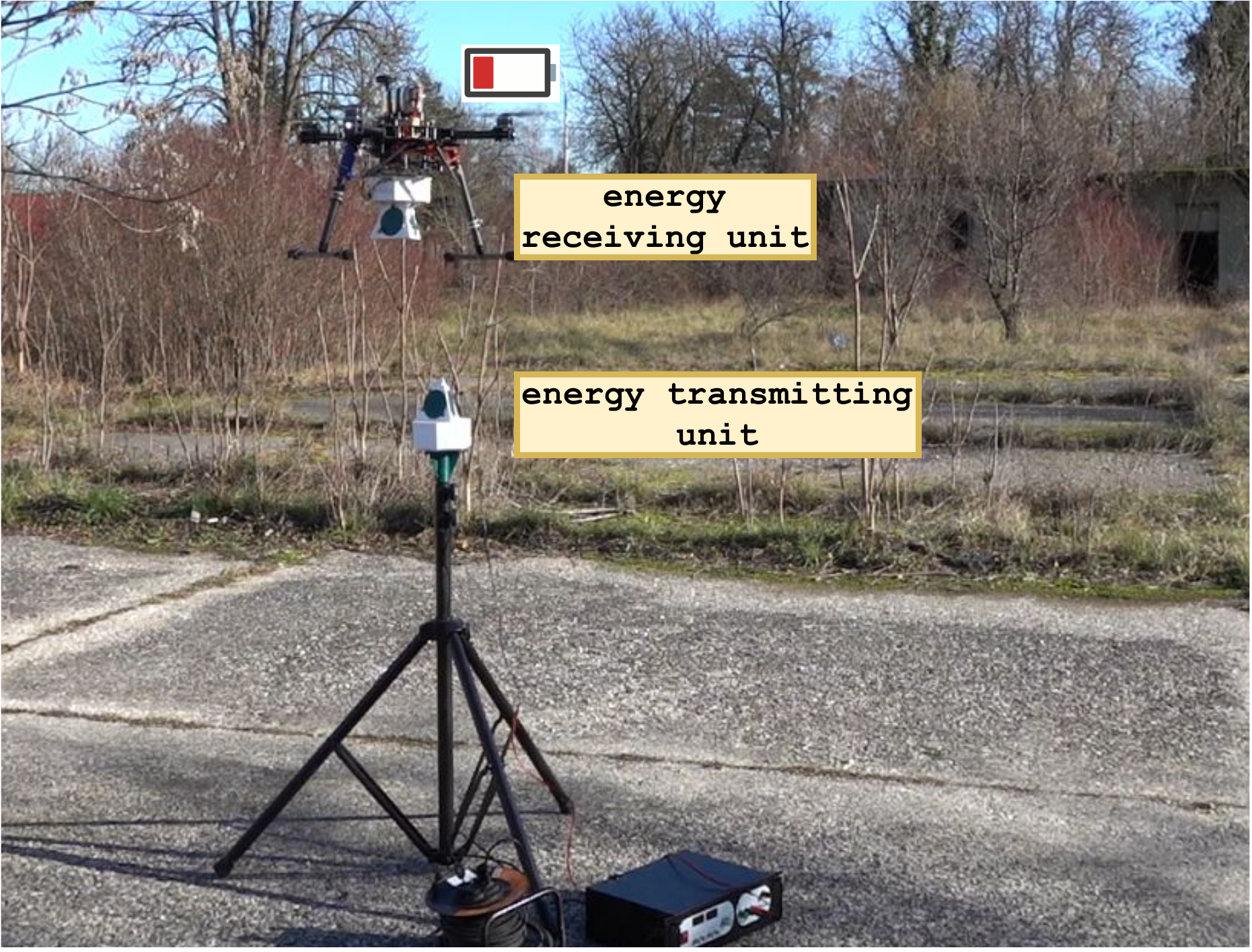}
\caption{Mid-flight landing sequence of the drone on the wireless docking station. The proposed system consists of two independent modules: the energy transmitting unit (attached to the camera tripod) and the energy receiving unit (attached to the drone).}
\label{fig:first_page_photo}
\vspace{-0.7cm}
\end{figure}

Estimated drone battery life today varies from 5 to 45 minutes for Li-Po batteries \cite{mourgelas2020}. In most cases, when the drone runs out of power, the battery is manually replaced. This approach is very time consuming, is not suitable for the increasingly common autonomous operation of drones, and requires a human operator on site. Therefore, two possible automatic/autonomous approaches to extend drone  power autonomy are being investigated and deployed: a) automatic/autonomous battery replacement (swapping) and b) automatic/autonomous battery recharging using recharging stations. In \cite{lee2015}, an autonomous battery swapping system for drones was developed, focusing on an accurate battery swapping mechanism and precise landing. However, the low robustness of the developed system has been reported, and there is no report on the scalability of their system to other types of drones. A big step further is presented in \cite{barrett2018}, where the drone's battery is replaced by a mobile manipulator. However, for all the appeal of battery swapping systems, they require high accuracy, precise positioning, and special design of batteries and battery compartments, making them far more complex systems than recharging stations \cite{silva2022}.

In \cite{ucgun2021}, a thorough overview of battery charging stations is given and divided into two groups: wired and wireless charging stations. Although the power transfer efficiency of wireless charging stations is lower, the reliability of such systems in terms of energy transfer and environmental changes is much higher than that of wired charging stations \cite{ucgun2021}. When designing the right charging station, a tradeoff between power transfer efficiency and reliability should be considered based on the application in which the drone will be used. For wireless charging stations for drones, one of the biggest challenges is dealing with potential misalignment of transmitter and receiver coils after landing. To this end, various positioning mechanisms are used in landing stations. In \cite{galimov2020}, a comprehensive literature review is provided on the classification and technical design of UAV positioning mechanisms in landing stations. The authors in \cite{galimov2020} classify drone positioning approaches in landing stations into two main groups: with and without positioning. Drone landing stations without positioning systems require a sophisticated drone control system to ensure the desired accuracy during landing. Once the drone is landed, the landing platform can hold the drone in that position. An example of this is presented in \cite{campi2021}, where the wireless drone charging station was designed for almost any landing position and orientation to achieve the highest efficiency of power transfer. However, using such a system outdoors is challenging because there is no protection for outdoor weather conditions. On the other hand, drone landing stations with positioning systems can be divided into two main groups: with active and passive positioning. Active positioning systems include actuators and mechanisms to adjust the position and orientation of the drone to the right degree. Such examples are very well presented in \cite{rohan2019}, \cite{choi2016}, where in both cases 2DoF landing platforms with active positioning systems for drones are developed to align the drone in the XY plane before wireless charging. Compared to passive positioning systems, these systems are more complex, less fail-safe, and require more time for the drone to reach the correct position. Passive positioning systems do not require the installation of actuators, mechanisms, and/or other moving objects. Such examples include funnels for one or for each leg, overhead funnels, and closed contours.

To underscore the importance of docking stations for drones, two major players in the drone market have released two commercial docking stations for drones in 2022: DJI dock\footnote{dji.com/si/dock} and Skydio dock\footnote{skydio.com/skydio-dock}. However, they are only supported for their specific drones and are not applicable or scalable for custom or other drones on the market or in research. For this reason, we have chosen to develop a universal, lightweight and affordable design of wireless drone docking station and associated drone charging station that can be used in a wide range of drones in both industrial and service sectors.

The main and original contribution of this paper is the design and validation of a novel wireless docking station for drones (see Fig. \ref{fig:first_page_photo}). The purpose of the design is twofold. First, the misalignment of receiver and transmitter modules is completely eliminated, and second, the pyramid-shaped docking station reduces the effects of ground effect during landing. In addition, we present the energy transmitting unit (ETU) and the corresponding energy receiving unit (ERU) as two independent modules. Also, to the best of our knowledge, we are also the first to use and implement three wireless charging modules to extend and maximize the power autonomy of the drone. With this in mind, two possible configurations of three-receiver modules can be achieved: 1. connected in series and 2. connected in parallel. Both configurations have been validated and tested during the experiments, and we claim that the power transfer efficiency of the wireless drone docking station with three wireless charging modules is greater when the receiver modules are connected in series.

Section II presents the design of the wireless drone docking station both in electrical (Section \ref{sec:design_electrical}) and mechanical (Section \ref{sec:design_mechanical}) aspects. In Section \ref{sec:validation} the validation of the proposed wireless drone docking station is conducted. For that, the power transfer efficiency is investigated (Section \ref{sec:validation_a}) and the drone landing experiment (Section \ref{sec:validation_drone}) is conducted. Finally, in Section  \ref{sec:conclusion} concluding remarks and future directions are discussed.
\section{System Design}
\label{sec:design}

In this section, the design of the wireless drone docking station is presented. In the first part, we describe the wireless charging module that was used in the system. Then, we present the electrical and mechanical design of the system in detail.

\subsection{Wireless Charging Module}
\label{sec:design_wireless}

\begin{figure}
\centering
\includegraphics[width=7cm]{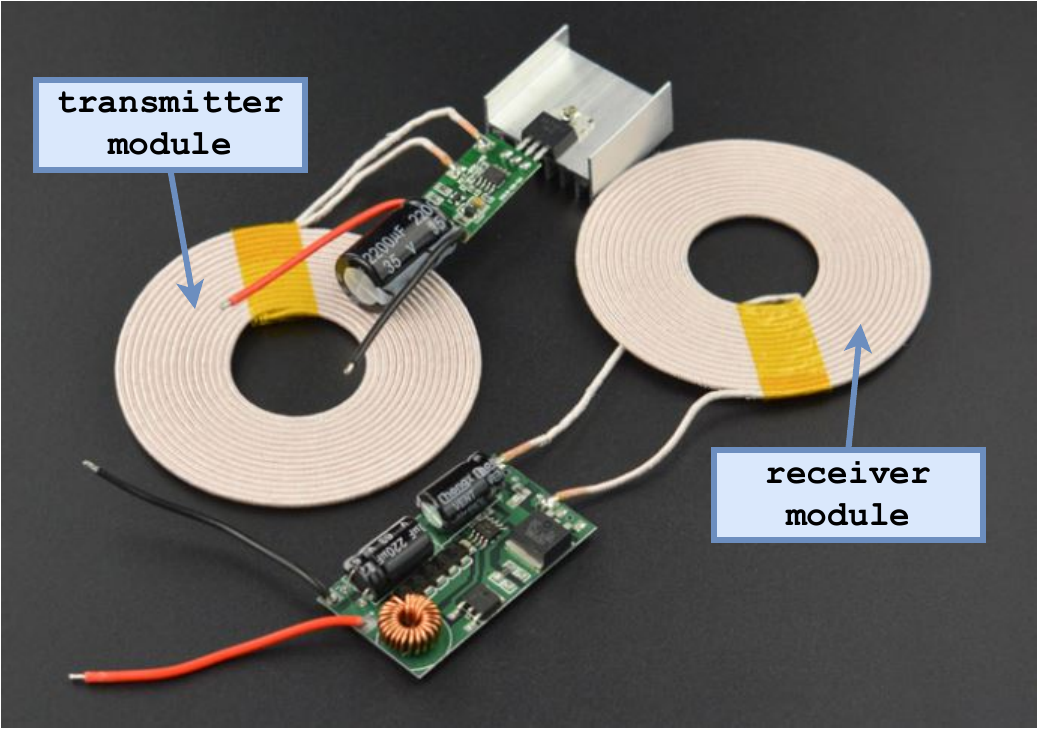}
\caption{\textit{Off-the-shelf} wireless charging module that consists of a transmitter and receiver module based on a magnetic resonant coupling wireless power transfer (WPT) technology \cite{wireless_charging_module}.}
\label{fig:wireless_charging_module}
\vspace{-0.7cm}
\end{figure}

The design and development of wireless charging modules is a well researched area. In the last few decades, much work has been put into the development and optimization of various wireless power transfer principles \cite{mohsan2022}, \cite{rong2022}, such as capacitive charging, inductive charging, magnetic resonance charging, etc. However, the focus of this work is not on the design, development or optimization of novel wireless charging modules. Therefore, we have taken proven \textit{off-the-shelf} wireless charging modules based on the specifications in Tab. \ref{tab:wpt_specs}, which are shown in Fig. \ref{fig:wireless_charging_module}, as it is demonstrated in \cite{jawad2022} and \cite{mohsan2022} that magnetic resonance coupling is the most effective wireless power transfer (WPT) technology. In addition, the maximum permissible distance (gap) between two coils to ensure maximum power transfer is up to 10 mm, and the efficiency of a pair of coils is specified as 95\%\cite{wireless_charging_module}.

\begin{table}
\centering
\caption{Technical specification of the \textit{off-the-shelf} wireless charging module \cite{wireless_charging_module}.}
\label{tab:wpt_specs}
\begin{tabular}{|c||c|}
\hline
input voltage $U_{in}$ & 24 VDC\\
\hline
output voltage $U_{out}$ & 12 VDC\\
\hline
output current $I_{out}$ & 3 A\\
\hline
coil size $\Phi d/\Phi D\mathrm{x}h$ & $\Phi30\mathrm{x}\Phi82\mathrm{x}2$ [mm]\\
\hline
efficiency $\eta$ & 95°\\
\hline
maximum permissible gap & 10 mm\\
\hline
wireless power transfer (WPT) type & magnetic resonant coupling\\
\hline
\end{tabular}
\end{table}

\subsection{Electrical Design}
\label{sec:design_electrical}

\begin{figure}
\centering
\includegraphics[width=8cm]{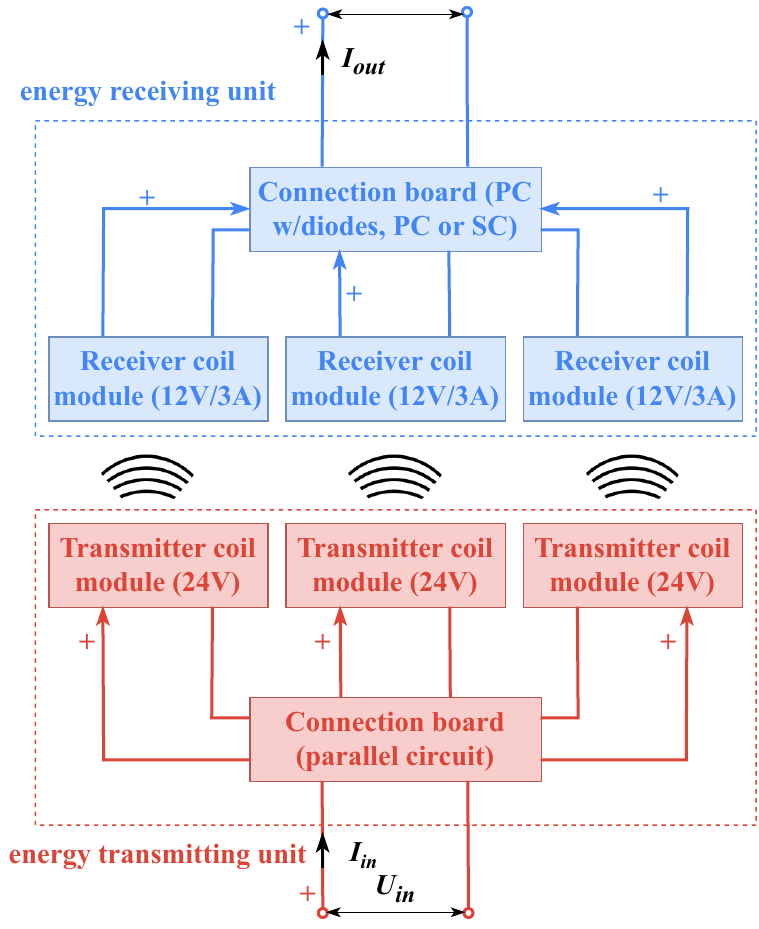}
\caption{Electrical scheme of the proposed wireless drone docking station, which consists of an ETU and an ERU. The system includes three wireless charging modules connected in parallel at the transmitter side. On the receiver side, the wireless charging modules can be connected in series or parallel.}
\label{fig:electrical_design}
\vspace{-0.3cm}
\end{figure}

\begin{figure}
\centering
\includegraphics[width=4.7cm]{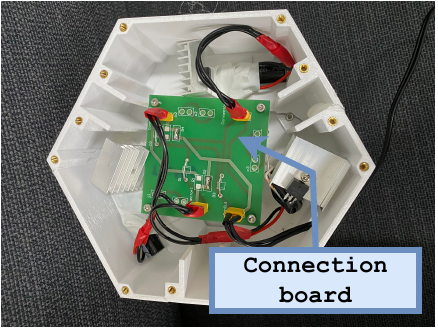}
\caption{The connection board placed and wired in the ETU. The same connection board is used in the ERU.}
\label{fig:board}
\vspace{-0.3cm}
\end{figure}

With the data from the wireless charging modules, we set out to design the electrical part of the wireless drone docking station. For this purpose, we divided the system into two modules: 1. ETU and 2. ERU. Since the main goal is wireless power transfer, there is no cable connection between the two modules, which makes them completely independent. To find out how exactly the three wireless charging modules behave in the system, we propose the following electrical scheme shown in Fig. \ref{fig:electrical_design}. On the ETU side, we connected three transmitter coil modules in parallel and supplied them with the input voltage $U_{in}$ specified by the manufacturer. For this purpose, we designed and fabricated a special printed circuit board (PCB), i.e. a connection board, that allows a more compact system (see Fig. \ref{fig:board}). On the ERU side, we created the possibility to connect three receiver coils in parallel or in series, depending on the output voltage $U_{out}$ we want to achieve on the drone side for charging the battery (12 V or 36 V). The same custom PCB design used in the transmitter is designed to be used in the receiver as well, allowing three receiver coils to be connected both in series and in parallel at the receiver side by simply switching jumpers.

\begin{figure}
\centering
\includegraphics[width=\linewidth]{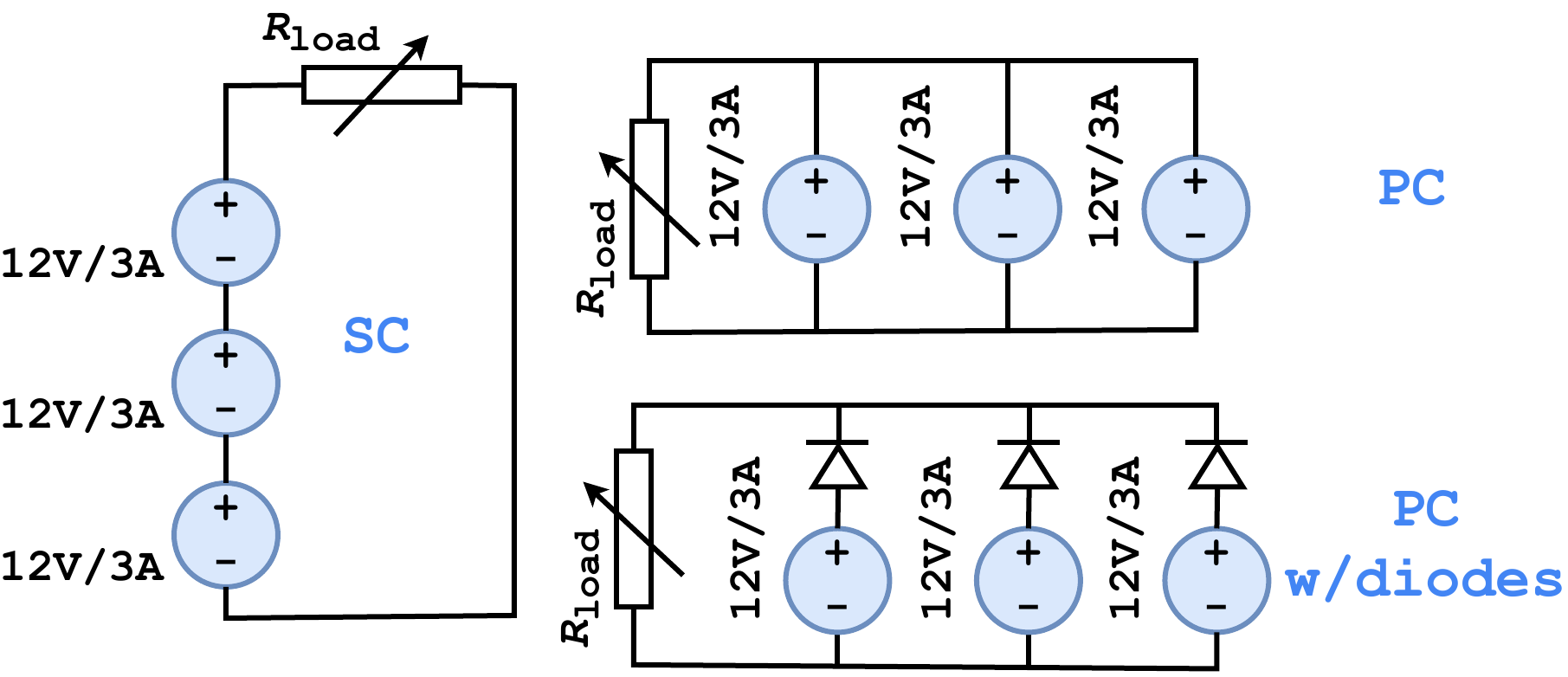}
\caption{Three methods of connecting receiver coils: series connection (SC), parallel connection with diodes (PC w/diodes), and parallel connection without diodes (PC).}
\label{fig:connecting}
\end{figure}

In designing the electrical system, we assumed that if the three receiver coils were connected in parallel, there could be potential challenges in the output current of each receiver coil that would affect the output of the other receiver coils by returning to that circuit. For this reason, we introduced the third connection method, with a diode between each coil and the common parallel connection. This method allows the power transfer system to remain functional even if one of the coils is shorted.  The three methods of connecting receiver coils are shown in Fig. \ref{fig:connecting}.

\subsection{Mechanical Design}
\label{sec:design_mechanical}

In our previous work \cite{stuhne2022}, we presented a design for a wireless drone recharging station and the associated special robot end effector to install the charging station on the power line. This research was part of the AERIAL CORE \footnote{aerial-core.eu} project, where one of the main goals is to enable autonomous drone inspections of large infrastructures. To this end, it is critical to enable and/or extend the power autonomy of drone without human interaction. However, the focus of the work presented in \cite{stuhne2022} was on the simple and reliable installation of the charging station with a robotic arm that will be attached to the drone in the future.

\begin{figure}
\centering
\includegraphics[width=0.8\linewidth]{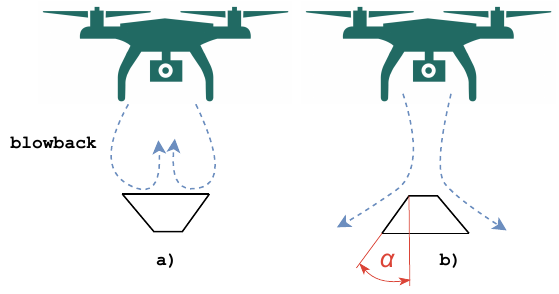}
\caption{Visualization of the pyramid-shaped landing pad. The tip of the pyramid-shaped landing pad can be oriented a) downward and b) upward, which significantly affects the occurrence of blowback (ground effect) during landing.}
\label{fig:v_landing}
\vspace{-0.3cm}
\end{figure}

Nevertheless, in \cite{stuhne2022}, the landing of the drone was also considered and a V-shaped landing pad was developed. The main purpose is to achieve passive positioning of the drone to avoid misalignment of the charging coils. During initial drone landing experiments, we encountered the blowback effect, which proved to be particularly significant and difficult. Initial experiments at the time indicated that simply rotating the tip of the V-landing pad 180 degrees would prevent the blowback effect because the airflow would not return to the drone (see Fig. \ref{fig:v_landing}).

This important finding that the tip of the V-landing pad can be rotated 180 degrees was the basis for the mechanical design of the wireless drone docking station. In \cite{obayashi2019}, the authors conceptually proposed the idea of a frustum-shaped 4-sided ERU, but did not implement it in practice. Moreover, it is questionable how feasible this idea is, since it is not discussed how the 4-sided frustum should enter the docking station. In our case, we decide to use the frustum-shaped docking station based on all the advantages presented in Section \ref{sec:intro}. However, certain design aspects such as the slope angle of the frustum and the number of frustum sides should be considered. To enable the passive positioning of the ERU to the ETU during landing, the following equation was considered according to \cite{galimov2020}:

\begin{equation}
\alpha \ge \arctan{(\mu)},
\label{eq:friction}
\end{equation}
where $\alpha$ is the slope angle of the frustum side and $\mu$ is the coefficient of friction between the contact surfaces of the ETU and the ERU. According to the Equation \eqref{eq:friction}, the landing and takeoff of the drone are secured to avoid self-locking. The main manufacturing process for producing design elements is FDM 3D printing (fused deposition modeling) and the main material is PETG (polyethylene terephthalate). The coefficient of friction $\mu$ is not accurate in this case and there is a lack of data on it, hence, the slope angle $\alpha$ was arbitrarily chosen to experimentally verify the landing of the drone. However, in any other case where the friction coefficient $\mu$ is known, Equation \eqref{eq:friction} should be applied.

The best frustum shape to ensure drone positioning regardless of drone orientation is the conical shape. However, the conical frustum is impractical for WPT because of the limited distance between the transmitter and receiver coils to achieve optimal power transfer efficiency. To ensure the smallest possible distance between two coils, we investigated pyramidal frustums. Since we are studying the system behavior with three wireless charging modules, we chose a three-sided pyramidal frustum. However, the orientation of the drone matters during landing, and a three-sided pyramidal frustum may cause a mismatch between the ERU and the ETU. To eliminate the possibility of the drone's orientation not matching the ETU's orientation during landing, we developed a separate part that we attached to the camera tripod that could rotate around the camera tripod clockwise and counterclockwise to provide passive rotational degree of freedom. The camera tripod served as a support for the ETU.

\begin{figure}
\centering
\includegraphics[width=0.9\linewidth]{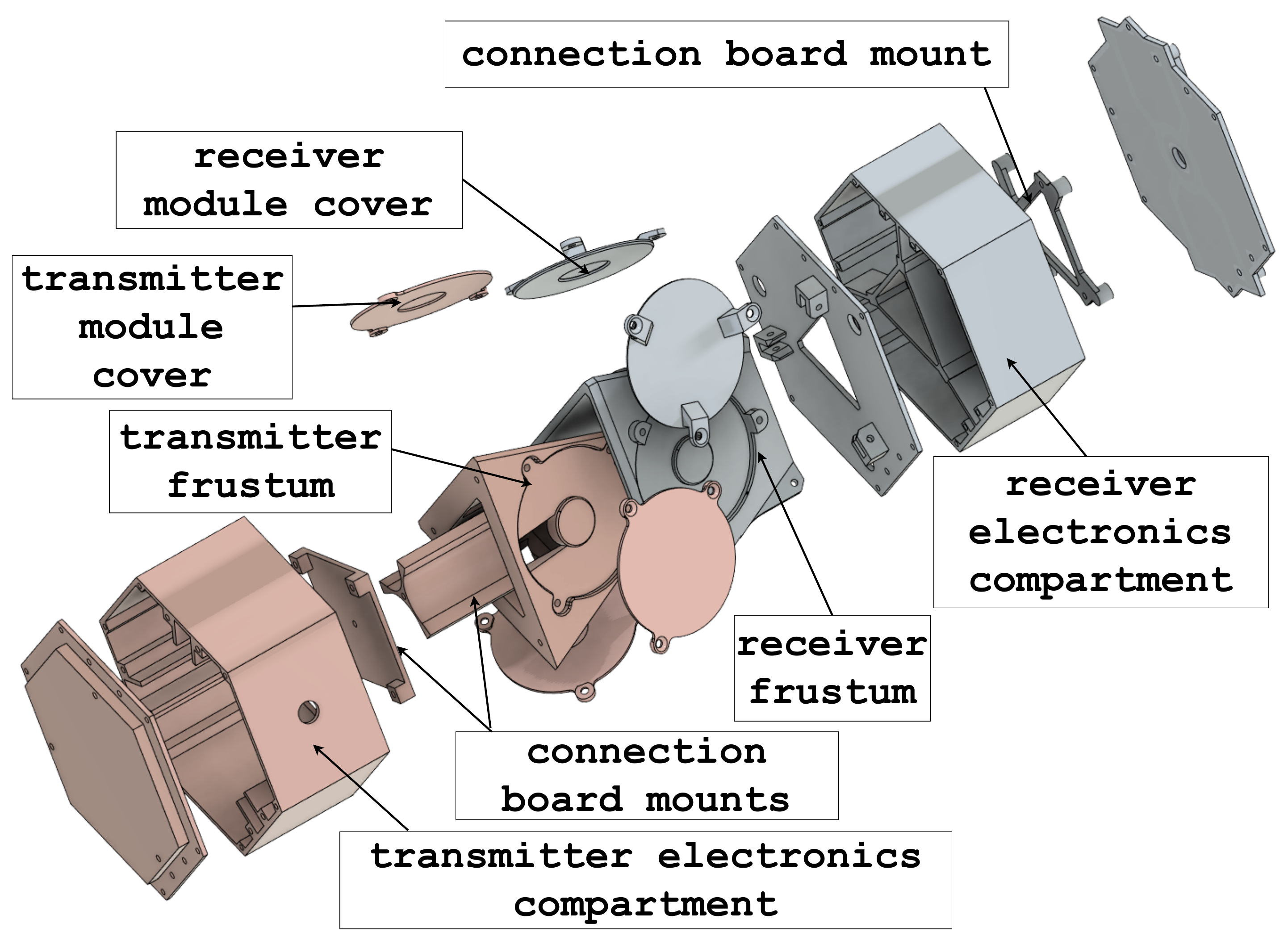}
\caption{Exploded view of the basic design elements of the ETU and the ERU in CAD.}
\label{fig:cad}
\end{figure}

Based on the geometric data of the wireless charging modules presented in Tab. \ref{tab:wpt_specs}, the geometric data of the wireless charging modules shown in Fig. \ref{fig:electrical_design}, the ETU and the ERU have been designed to have all the necessary elements (receiver and transmitter coils, receiver and transmitter modules, connection boards, etc.). All these elements are properly mounted and placed in the ETU and ERU, and the corresponding design elements and groups are shown in Fig. \ref{fig:cad}.

\section{System Validation}
\label{sec:validation}

In this section, firstly we validate the electrical design of the system by verifying the wireless power transfer efficiency $\eta$ and the output voltage $P_{out}$ for series and parallel connections, and secondly, we also verify the mechanical design with the drone landing and take off experiment.

\subsection{Charging Validation}
\label{sec:validation_a}

\begin{figure}
\centering
\includegraphics[width=0.5\linewidth]{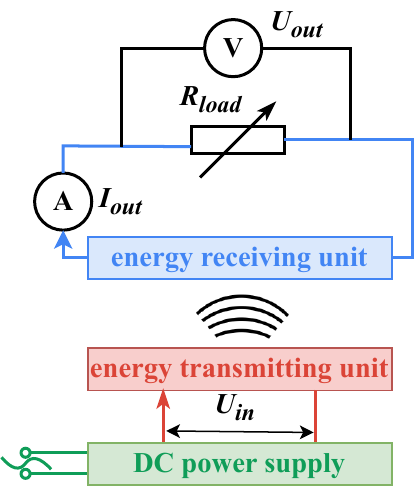}
\caption{Schematic diagram of the experimental setup used for the charging performance experiment. The ETU was connected to the DC power supply with a constant input voltage $U_{in}$ of 24 V. A variable resistor was connected to the output of the ERU to test the series and parallel connections of the receiving side under different loads. During the experiment, the output voltage $U_{out}$ and current $I_{out}$ were measured.}
\label{fig:setup_block}
\vspace{-0.7cm}
\end{figure}

After the design process was completed, we conducted the experiment on charging performance. For this purpose, the following experimental setup was designed (Fig. \ref{fig:setup_block}). The ETU was connected to the DC power supply with an input voltage $U_{in}$ of 24 V, which is prescribed by the manufacturer of the wireless charging modules. To the ERU, we connected a variable resistor with resistance $R_{load}$ to avoid overcurrent, and to simulate load. In this way, we connected an ammeter and a voltmeter to measure the current $I_{out}$ and output voltage $U_{out}$, respectively. The resistance range of a variable resistor was crucial for this experiment and was chosen differently for each configuration based on the theoretical output voltage at the receiving side. The resistance range of a variable resistor was chosen as follows:

\begin{equation}
  R_{load} =
    \begin{cases}
      1-10\;\mathrm{\Omega,} & \text{for parallel circuit,}\\
      10-100\;\mathrm{\Omega,} & \text{for series circuit,}\\
    \end{cases}
\label{eq:Rload}
\end{equation}
to accompany the theoretical output voltage, which is described as follows:

\begin{equation}
  U_{out} =
    \begin{cases}
      12\;\mathrm{V,} & \text{for parallel circuit w/diodes,}\\
      12\;\mathrm{V,} & \text{for parallel circuit,}\\
      36\;\mathrm{V,} & \text{for series circuit.}\\
    \end{cases}
\label{eq:Uout}
\end{equation}

Because the theoretical output voltage of each configuration is predetermined, an appropriate load resistor should be applied so as not to exceed the maximum allowable output current in each configuration ($I_{SCout}=3\:\mathrm{A}$ and $I_{PCout}=9\:\mathrm{A}$). In each experiment, we slightly changed the load resistance at the output. Then the output voltage and current were measured, and the input current $I_{in}$ was read from the DC power supply. Finally, when the data were collected, we calculated the transmitted output power $P_{out}$ for each experiment as follows:

\begin{equation}
P_{out} = U_{out}I_{out},
\label{eq:output_power}
\end{equation}
and the power transfer efficiency $\eta$ for each experiment as:

\begin{equation}
\eta = \frac{U_{out}I_{out}}{U_{in}I_{in}}.
\label{eq:efficiency}
\end{equation}

\begin{figure}
\centering
\includegraphics[width=\linewidth]{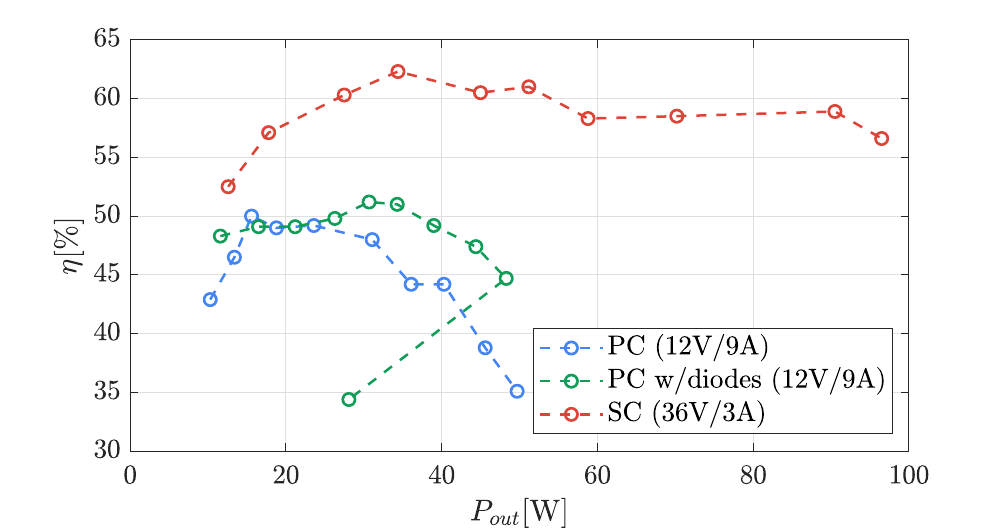}
\caption{The power transfer efficiency of the wireless drone docking station when the receiver side is connected in series and parallel. The losses are larger in the PC configuration (parallel connection) than in the SC configuration (series connection) due to the larger output current.}
\label{fig:efficiency}
\vspace{-0.3cm}
\end{figure}

\begin{figure}
\centering
\begin{subfigure}{0.45\textwidth}
\includegraphics[width=\linewidth]{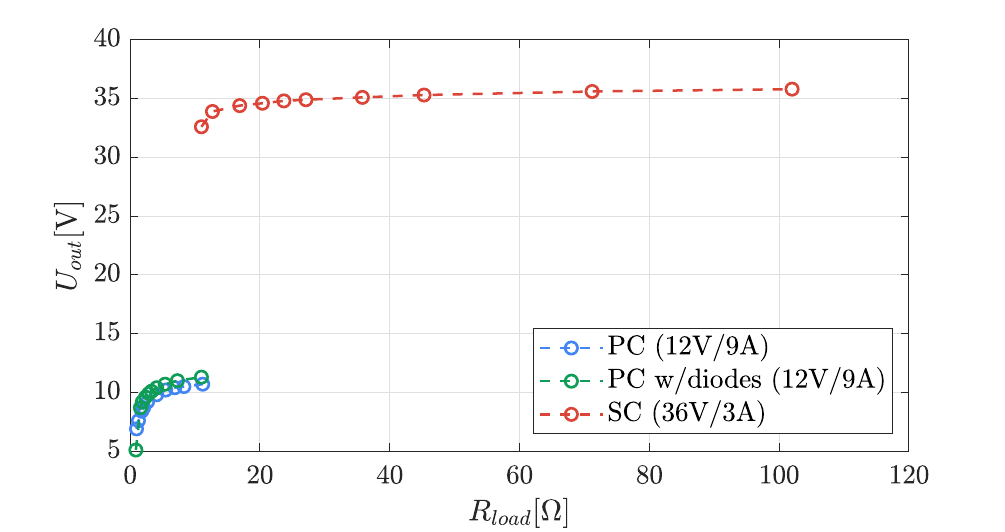}
    \caption{}
    \label{fig:voltage_all}
\end{subfigure}
\hfill
\begin{subfigure}{0.45\textwidth} \includegraphics[width=\linewidth]{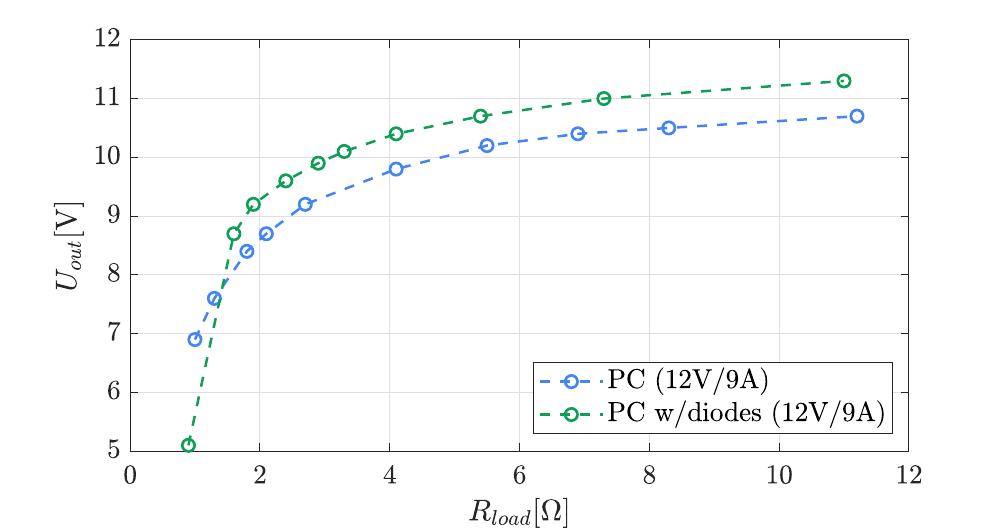}
    \caption{}
    \label{fig:voltage_parallel}
\end{subfigure}
\caption{Dependence of output voltage $U_{out}$ on load change $R_{load}$ for three wireless charging modules in series and parallel connection.}
\label{fig:voltage}
\vspace{-0.7cm}
\end{figure}

The graphical representation of the power transfer efficiency for each configuration is shown in Fig. \ref{fig:efficiency}. The output current is higher for both PC configurations (with and without diodes), resulting in larger losses. Each configuration reaches some optimum at some point, but the power transfer efficiency range is higher and more stable for the SC configuration. An interesting situation occurs in the PC configuration with diodes, where there is a sudden drop in power transfer efficiency and output power. In this case, the output voltage at the receiver module drops sharply (see Fig. \ref{fig:voltage_parallel}), which is due to large voltage drops across the resistors and diodes. A large voltage drop across resistors and diodes means losses in the circuit caused by large currents.

There is a significant difference in power transfer efficiency between series and parallel connections. We report a maximum output power of 96.5 W at the power transfer efficiency of 56.6 \% for the series connection of receiver coils, which is almost twice the power reported for parallel connections with and without diodes (Fig. \ref{fig:efficiency}). With the obvious advantage of higher power transfer efficiency and higher power transfer, the output voltage in the series connection is more stable with only a small drop at higher loads (Fig. \ref{fig:voltage}). In parallel circuits with and without diodes, the drop in output voltage is more pronounced, especially at higher output currents. The probable reason for the higher power transfer efficiency, higher output power, and more stable output voltage in the series connection is the three times smaller nominal output current compared to the parallel connections, which means that the losses are lower ($I_{outSeries}=3\mathrm{~A}$ and $I_{outParallel}=9\mathrm{~A}$).

\subsection{Drone Landing and Take Off}
\label{sec:validation_drone}

\begin{figure}
\centering
\begin{subfigure}{0.2\textwidth}
\centering
\includegraphics[width=\linewidth]{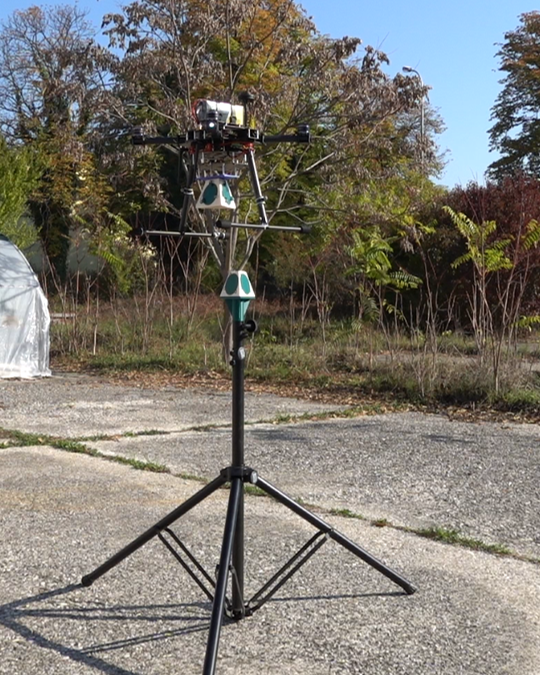}
    \caption{}
    \label{fig:first}
\end{subfigure}
\hfill
\begin{subfigure}{0.2\textwidth}
\centering
\includegraphics[width=\linewidth]{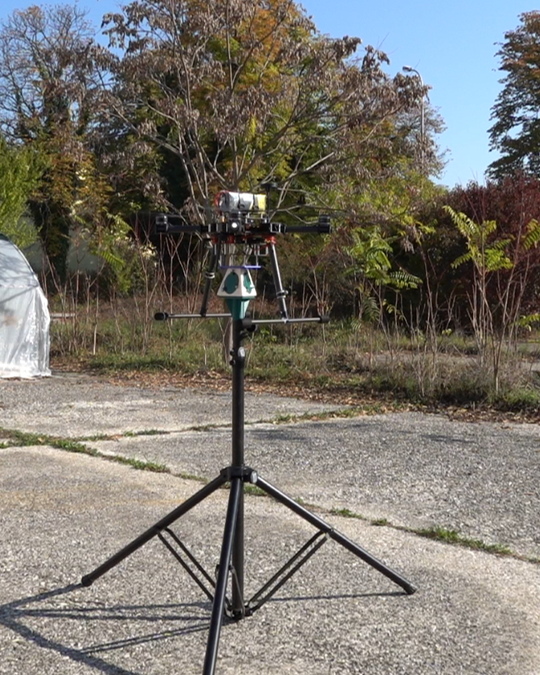}
    \caption{}
    \label{fig:third}
\end{subfigure}
\caption{The drone landing test to verify the slope angle. The drone landing and take off sequence were successfully tested several times, which experimentally demonstrated the accuracy of the slope angle.}
\label{fig:setup_real}
\vspace{-0.3cm}
\end{figure}

To verify the accuracy of the slope angle $\alpha$, we designed an experiment in which a drone manually lands on the ETU (see Fig. \ref{fig:setup_real}). For this purpose, we attached the ERU to the drone. We also attached the ETU to the camera tripod with the additional adapter. In this way, the rotational degree of freedom of the ETU adapter around the camera tripod was enabled since both adapter and the end of the camera tripod have cylindrical surface. The drone successfully landed and took off 
in a series of repeated experiments, 
which was the proof of choosing the right slope angle $\alpha$. For more details about the drone landing, please see the supplemental video. The specifications of the wireless drone docking station are given in the Tab. \ref{tab:specs}.

\begin{table}
\centering
\caption{Technical specification of the wireless drone docking station.}
\label{tab:specs}
\begin{tabular}{|c||c|}
\hline
ETU mass $m_{ETU}$ & 700 g\\
\hline
ERU mass $m_{ERU}$ & 700 g\\
\hline
ETU size $b\mathrm{x}h\mathrm{x}l$ &  178x155x178[mm]\\
\hline
ERU size $b\mathrm{x}h\mathrm{x}l$ & 178x155x179 [mm]\\
\hline
slope angle $\alpha$ & 12.7°\\
\hline
\end{tabular}
\vspace{-0.3cm}
\end{table}

\section{Conclusion}
\label{sec:conclusion} 

In this paper, a novel wireless drone docking station was presented. Three wireless charging modules were coupled together to extend the power autonomy of the drone on the pyramid-shaped docking station, which enables automatic alignment of the transmitter and receiver charging coils. We tested both the landing and charging capabilities of the proposed docking station with the application on the drone. The technology presented in this paper is ideally suited to extend the power autonomy of any drone or drone swarm because it is easy to implement, universal, and scalable.

The first step of our future work will be the drone landing experiment, where we will additionally charge the drone battery with the energy transmitted wirelessly by the ETU. The internal resistance of the drone battery and the battery management system (BMS) will be taken into account. The BMS will ensure proper charging on the ERU side, while the internal resistance is of utmost importance as it changes during the charging process. The next step would then be to implement and adapt the developed wireless drone docking station to the wireless charging station presented in \cite{stuhne2022}. In addition, autonomous landing of drones with visual or magnetic feedback will be studied in detail.

\addtolength{\textheight}{-12cm}

\section*{ACKNOWLEDGMENT}
We thank our colleagues Jurica Goričanec and Antun Ivanović for their support and helpful insights during the drone landing and take-off trials.

\bibliographystyle{ieeetr}
\bibliography{bibliography}

\begin{thebibliography}{10}

\bibitem{statistaGlobalDrone}
``{G}lobal drone market by segment 2020-2030 | {S}tatista --- statista.com.'' \url{statista.com/statistics/1234538/worldwide-drone-market-by-segment}.
\newblock [Accessed 31-Jan-2023].

\bibitem{mourgelas2020}
C.~Mourgelas, S.~Kokkinos, A.~Milidonis, and I.~Voyiatzis, ``Autonomous drone charging stations: A survey,'' 2020.

\bibitem{lee2015}
D.~Lee, J.~Zhou, and W.~T. Lin, ``Autonomous battery swapping system for quadcopter,'' in {\em 2015 International Conference on Unmanned Aircraft Systems (ICUAS)}, pp.~118--124, June 2015.

\bibitem{barrett2018}
E.~Barrett, M.~Reiling, S.~Mirhassani, R.~Meijering, J.~Jager, N.~Mimmo, F.~Callegati, L.~Marconi, R.~Carloni, and S.~Stramigioli, ``Autonomous battery exchange of uavs with a mobile ground base,'' in {\em 2018 IEEE International Conference on Robotics and Automation (ICRA)}, pp.~699--705, 2018.

\bibitem{silva2022}
S.~C. De~Silva, M.~Phlernjai, S.~Rianmora, and P.~Ratsamee, ``Inverted docking station: A conceptual design for a battery-swapping platform for quadrotor uavs,'' {\em Drones}, vol.~6, no.~3, 2022.

\bibitem{ucgun2021}
H.~Ucgun, U.~Yuzgec, and C.~Bayilmis, ``A review on applications of rotary-wing unmanned aerial vehicle charging stations,'' {\em International Journal of Advanced Robotic Systems}, vol.~18, 2021.

\bibitem{galimov2020}
M.~Galimov, R.~Fedorenko, and A.~Klimchik, ``Uav positioning mechanisms in landing stations: Classification and engineering design review,'' {\em Sensors}, vol.~20, no.~13, 2020.

\bibitem{campi2021}
T.~Campi, S.~Cruciani, F.~Maradei, and M.~Feliziani, ``Efficient wireless drone charging pad for any landing position and orientation,'' {\em Energies}, vol.~14, 2021.

\bibitem{rohan2019}
A.~Rohan, M.~Rabah, F.~Asghar, M.~Talha, and S.~H. Kim, ``Advanced drone battery charging system,'' {\em Journal of Electrical Engineering and Technology}, vol.~14, 2019.

\bibitem{choi2016}
C.~H. Choi, H.~J. Jang, S.~G. Lim, H.~C. Lim, S.~H. Cho, and I.~Gaponov, ``Automatic wireless drone charging station creating essential environment for continuous drone operation,'' in {\em 2016 International Conference on Control, Automation and Information Sciences (ICCAIS)}, pp.~132--136, 2016.

\bibitem{mohsan2022}
S.~A.~H. Mohsan, N.~Q.~H. Othman, M.~A. Khan, H.~Amjad, and J.~Żywiołek, ``A comprehensive review of micro uav charging techniques,'' {\em Micromachines}, vol.~13, no.~6, 2022.

\bibitem{rong2022}
C.~Rong, X.~He, Y.~Wu, Y.~Qi, R.~Wang, Y.~Sun, and M.~Liu, ``Optimization design of resonance coils with high misalignment tolerance for drone wireless charging based on genetic algorithm,'' {\em IEEE Transactions on Industry Applications}, vol.~58, pp.~1242--1253, Jan 2022.

\bibitem{jawad2022}
A.~M. Jawad, R.~Nordin, H.~M. Jawad, S.~K. Gharghan, A.~Abu-Samah, M.~J. Abu-Alshaeer, and N.~F. Abdullah, ``Wireless drone charging station using class-e power amplifier in vertical alignment and lateral misalignment conditions,'' {\em Energies}, vol.~15, 2 2022.

\bibitem{wireless_charging_module}
DFRobot, ``{Wireless Charging Module 12V/3A}.'' \url{dfrobot.com/product-2087.html}, note = {Accessed: 2023-01-28}.

\bibitem{stuhne2022}
D.~Stuhne, V.~D. Hoang, G.~Vasiljevic, S.~Bogdan, Z.~Kovacic, A.~Ollero, and E.~S.~M. Ebeid, ``Design of a wireless drone recharging station and a special robot end effector for installation on a power line,'' {\em IEEE Access}, vol.~10, pp.~88719--88737, 2022.

\bibitem{obayashi2019}
S.~Obayashi, Y.~Kanekiyo, K.~Nishizawa, and H.~Kusada, ``85-khz band 450-w inductive power transfer for unmanned aerial vehicle wireless charging port,'' in {\em 2019 IEEE Wireless Power Transfer Conference (WPTC)}, pp.~80--84, 2019.

\end{thebibliography}

\end{document}